\documentclass[10pt,twocolumn,letterpaper]{article}

\newcommand{\Tref}[1]{Table~\ref{#1}}
\newcommand{\eref}[1]{Eq.~(\ref{#1})}

\newcommand{\fref}[1]{Fig.~\ref{#1}}

\newcommand{\sref}[1]{Sec.~\ref{#1}}

\usepackage{iccv}
\usepackage{times}
\usepackage{epsfig}
\usepackage{graphicx}
\usepackage{amsmath}
\usepackage{amssymb}

\usepackage{enumitem}
\usepackage{xfrac}
\usepackage{ctable}
\usepackage{array}
\usepackage{caption}
\usepackage{multirow, makecell}
\usepackage{subfig}
\usepackage{tabularx}

\renewcommand{\paragraph}[1]{\vspace{1mm}\noindent\textbf{#1}}

\newcolumntype{P}[1]{>{\centering\arraybackslash}p{#1}}
\newcolumntype{M}[1]{>{\centering\arraybackslash}m{#1}}
\newcolumntype{?}[1]{!{\vrule width #1}}

\newlength{\Oldarrayrulewidth}

\usepackage[pagebackref=true,breaklinks=true,letterpaper=true,colorlinks,bookmarks=false]{hyperref}

\iccvfinalcopy 

\def\iccvPaperID{2838} 

\ificcvfinal\fi
\begin{document}

\title{Video Object Segmentation using Space-Time Memory Networks}

\author{
Seoung Wug Oh\thanks{This work was done during an internship at Adobe Research.}\\Yonsei University \and Joon-Young Lee\\Adobe Research \and  Ning Xu\\Adobe Research \and  Seon Joo Kim\\Yonsei University
}

\maketitle
\thispagestyle{empty}

\begin{abstract}
We propose a novel solution for semi-supervised video object segmentation. 
By the nature of the problem, available cues (\eg video frame(s) with object masks) become richer with the intermediate predictions.
However, the existing methods are unable to fully exploit this rich source of information.
We resolve the issue by leveraging memory networks and learn to read relevant information from all available sources. 
In our framework, the past frames with object masks form an external memory, and the current frame as the query is segmented using the mask information in the memory.
Specifically, the query and the memory are densely matched in the feature space, covering all the space-time pixel locations in a feed-forward fashion.
Contrast to the previous approaches, the abundant use of the guidance information allows us to better handle the challenges such as appearance changes and occlussions.
We validate our method on the latest benchmark sets and achieved the  
state-of-the-art performance (overall score of 79.4 on Youtube-VOS val set, $\mathcal{J}$ of 88.7 and 79.2 on DAVIS 2016/2017 val set respectively) while having a fast runtime (0.16 second/frame on DAVIS 2016 val set).

\end{abstract}

\section{Introduction}
Video object segmentation is a task of separating the foreground and the background pixels in all frames of a given video. It is an essential step for many video editing tasks, which is getting more attention as videos have become the most popular form of shared media contents.
We tackle the video object segmentation problem in the semi-supervised setting, where the ground truth mask of the target object is given in the first frame and the goal is to estimate the object masks in all other frames.
It is a very challenging task as the appearance of the target object can change drastically over time and also due to occlusions and drifts. 

\begin{figure}
\centering
\includegraphics[width=1.0\linewidth]{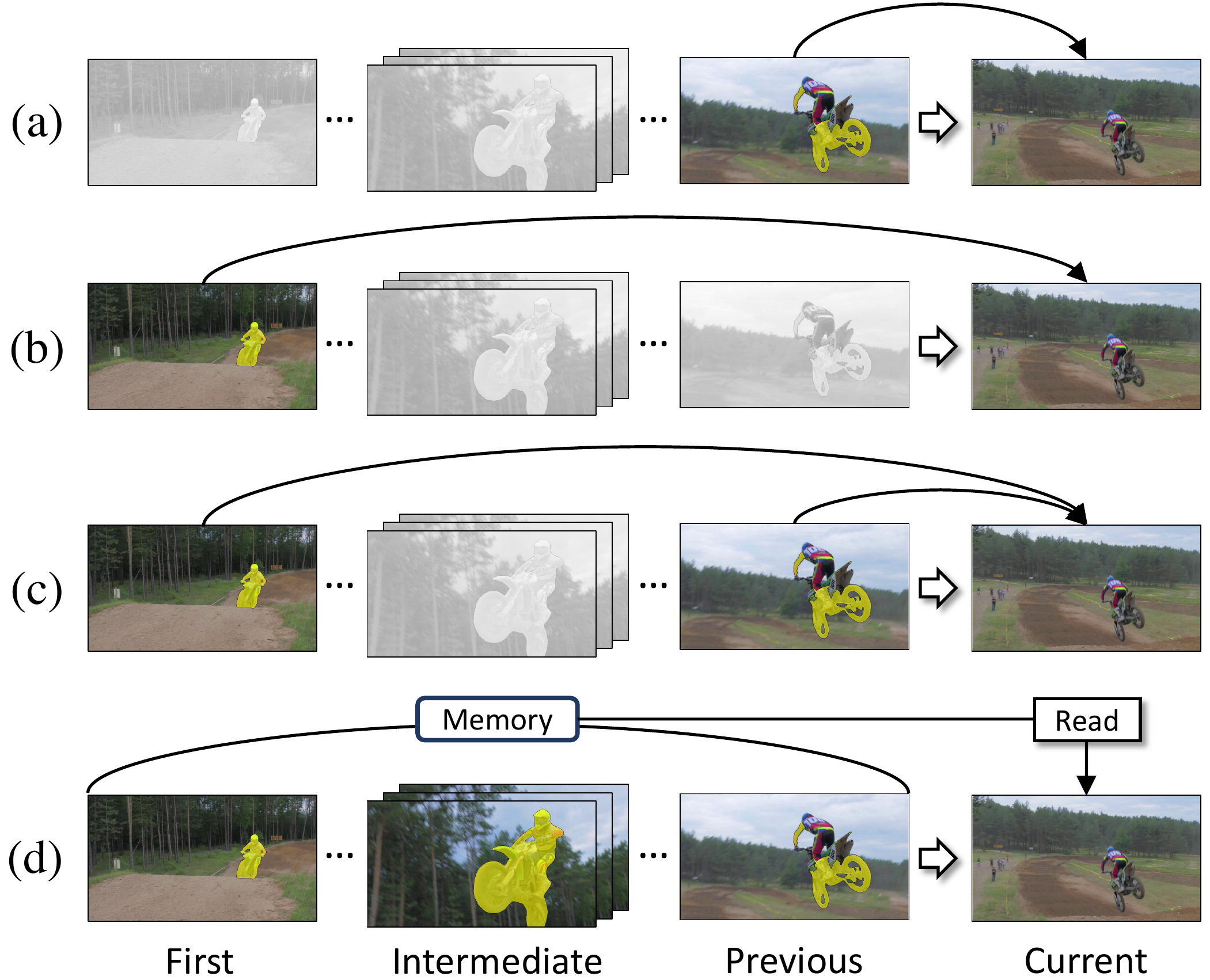}
\caption{
Previous DNN-based algorithms extract features in different frames for video object segmentation (a-c). We propose an efficient algorithm that exploits multiple frames in the given video for more accurate segmentation (d).
}
\label{Fig:teaser}
\end{figure}

As in most tasks in computer vision, many deep learning based algorithms have been introduced to solve the video object segmentation problem. 
With deep learning approaches, the essential question is from which frame(s) should the deep networks learn the cues?
In some algorithms, the features were extracted and propagated from the previous frame (\fref{Fig:teaser}(a))~\cite{hu2017maskrnn, perazzi2017learning}.
The main strength of this approach is that it can deal with changes in appearance better, while sacrificing robustness against occlusions and error drifts. 
Another direction for deep learning based video segmentation is to use the first frame as a reference and independently detect the target object at each frame (\fref{Fig:teaser}(b))~\cite{caelles2017one, yoon2017pixel, hu2018videomatch}. The pros and cons of this approach are exactly the opposite from the previous approach. 
Methods that use both the first frame and the previous frame to take the advantages of the two approaches were proposed in~\cite{oh2018fast,yang2018efficient} (\fref{Fig:teaser}(c)). By using two frames as the source for cues, the algorithm~\cite{oh2018fast} achieved the state-of-the-art accuracy with faster running time, as the algorithm does not require online learning as with other methods. 

As using two frames has shown to be beneficial for video segmentation, a natural extension is to use more frames, possibly every frame in the video, for the segmentation task. 
The question is how to design an efficient deep neural network (DNN) architecture that exploits all the frames. 
In this paper, we propose a novel DNN system based on the memory network~\cite{sukhbaatar2015end,miller2016key,kumar2016ask} that computes the spatio-temporal attention on every pixel in multiple frames of the video for each pixel in the query image, to decide whether the pixel belongs to a foreground object or not.
With our framework, there is no restriction on the number of frames to use and new information can be easily added by putting them onto the memory.
This memory update greatly helps us to address the challenges like appearance changes and occlussions with no cost.
In addition to using more temporal information, our network inherently includes non-local spatial pixel matching mechanism that is well suited for pixel-level estimation problems. 
By exploiting rich reference information, our approach can deal with appearance changes, occlusions, and drifts much better than the previous methods.
Experimental results show that our method outperforms all the existing methods on public benchmark datasets by a large margin in terms of both speed and accuracy.

\section{Related Work}
\subsection{Semi-supervised Video Object Segmentation}
\paragraph{Propagation-based methods}~\cite{perazzi2017learning, khoreva2017lucid, hu2017maskrnn, li2018video} learn an object mask propagator, a deep network that refines misaligned mask toward the target object (\fref{Fig:teaser}(a)). 
To make the network object-specific, online training data is generated from the first frame by deforming the object mask~\cite{perazzi2017learning} or synthesizing images~\cite{khoreva2017lucid} for fine-tuning.
Li~\etal~\cite{li2018video} integrate re-identification module into the system to retrieve missing objects due to drifts.

\paragraph{Detection-based methods}~\cite{caelles2017one, maninis2017video, yoon2017pixel, bao2018cnn, chen2018blazingly, hu2018videomatch} work by learning an object detector using the object appearance on the first frame (\fref{Fig:teaser}(b)).
In \cite{caelles2017one, maninis2017video}, an object-specific detector learned by fine-tuning the deep networks at the test time is used to segment out the target object. 
In~\cite{chen2018blazingly, hu2018videomatch}, to avoid the online learning, pixels are embedded into feature space and classified by matching to templates.

\paragraph{Hybrid methods}~\cite{yang2018efficient, oh2018fast} are designed to take advantages of both detection and propagation approaches (\fref{Fig:teaser}(c)).
In~\cite{oh2018fast,yang2018efficient}, networks that exploit both the visual guidance from the first frame and the spatial priors from the previous frame were proposed.
Furthermore, some methods tried to exploit all previous information~\cite{xu2018youtube, voigtlaender2017online}. 
In~\cite{xu2018youtube}, a sequence-to-sequence network that learns the long-term information in videos was proposed.
Voigtlaender and Leibe~\cite{voigtlaender2017online} employ the idea of online adaptation and continuously update the detector using the intermediate outputs.

\paragraph{Online/Offline learning.} 
Many of aforementioned methods fine-tune deep network models on the initial object mask in the first frame to remember the appearance of the target object~\cite{caelles2017one,voigtlaender2017online,perazzi2017learning, khoreva2017lucid, perazzi2017learning, hu2017maskrnn, li2018video} during the test time. 
While the online learning improves accuracy, it is computationally expensive, limiting its practical use.
Offline learning methods attempted to bypass the online learning while retaining the accuracy~\cite{oh2018fast, yang2018efficient, chen2018blazingly, hu2018videomatch, joakim2018generative, rvos2019, feelvos2019}. 
A common idea is to design deep networks capable of object-agnostic segmentation at the test time, given guidance information.  

Our framework belongs to the offline learning method. 
Our framework maintains intermediate outputs in the external memory rather than fixing which frame(s) to use as the guidance, and adaptively selects necessary information in runtime.
This flexible use of the guidance information makes our method to outperform the aforementioned methods by a large margin.
Our memory network is also fast, as the memory reading is done as a part of the network forward pass, thus no online learning is required.
\begin{figure*}
\centering
\includegraphics[width=1.0\linewidth]{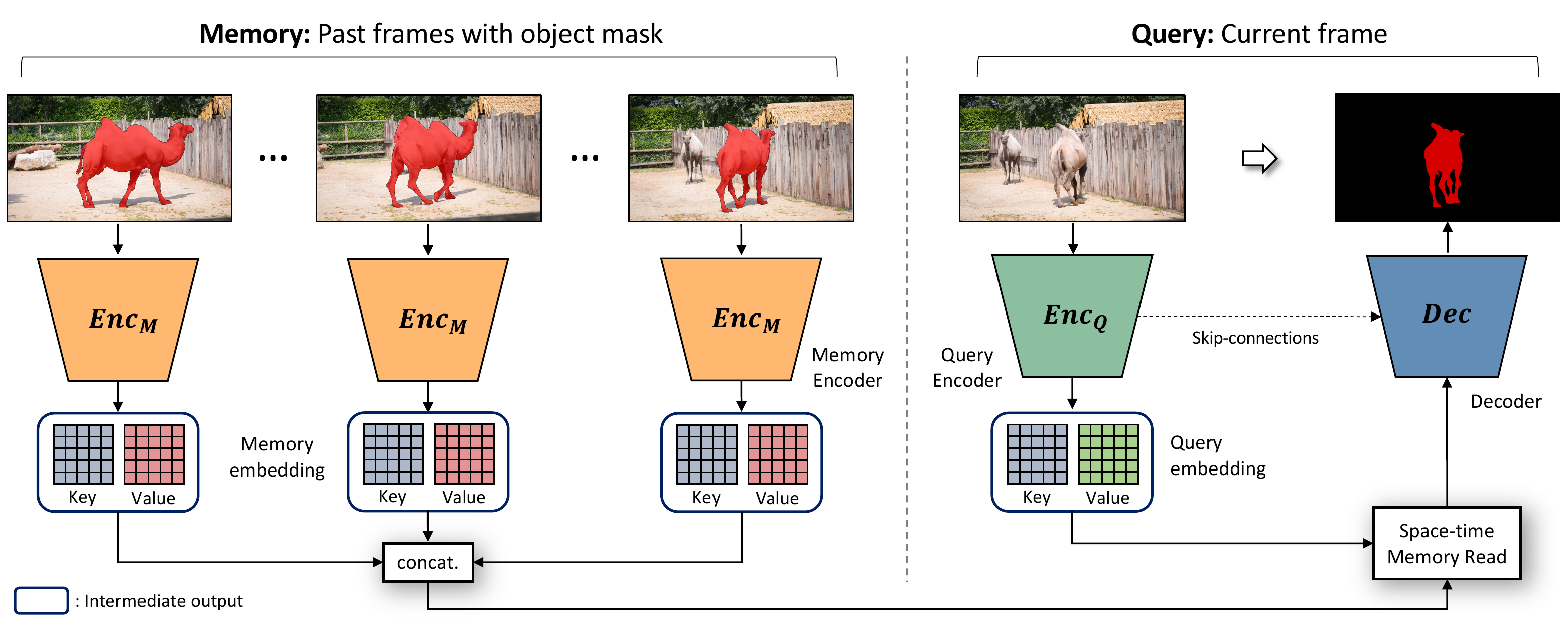}
\caption{Overview of our framework. Our network consists of two encoders each for the memory and the query frame, a space-time memory read block, and a decoder. The memory encoder ($Enc_M$) takes an RGB frame and the object mask. The object mask is represented as a probability map (the softmax output is used for estimated object masks). The query encoder ($Enc_Q$) takes the query image as input.}
\label{Fig:networks}
\end{figure*}

\subsection{Memory Networks}
Memory networks refer to the neural networks that have external memory where information can be written and read by purposes.
Memory networks that can be trained end-to-end were first proposed in the NLP research for the purpose of document Q\&A~\cite{sukhbaatar2015end, miller2016key,kumar2016ask}.
Commonly in those approaches, memorable information is separately embedded into key (input) and value (output) feature vectors.
Keys are used to address relevant memories whose corresponding values are returned. 
Recently, the memory networks have been applied to some vision problems such as personalized image captioning~\cite{park2017attend}, visual tracking~\cite{yang2018learning}, movie understanding~\cite{na2017read}, and summarization\cite{lee2018memory}.

While our work is based on the memory networks, we extend the idea of the memory networks to make it suitable for our task, semi-supervised video object segmentation. 
Obviously, frames with object masks are put to the memory, and a frame to be segmented acts as the query. 
The memory is dynamically updated with newly predicted masks and it greatly helps us to address the challenges like appearance changes, occlusions, and error accumulations without the online learning.

Our goal is to have pixel-wise predictions given a set of annotated frame(s) as memory.
Thus each pixel in the query frame needs to access information in the memory frames at different space-time locations.
To this end, we coin our memory into 4D tensors to contain pixel-level information and propose the space-time memory read operation to localize and read relevant information from the 4D memory. 
Conceptually, our memory reading can be considered as a spatio-temporal attention algorithm because we are computing \textit{when-and-where} to attend for each query pixel to decide whether the pixel belongs to a foreground object or not.

\section{Space-Time Memory Networks (STM)}
In our framework, video frames are sequentially processed starting from the second frame using the ground truth annotation given in the first frame.
During the video processing, we consider the past frames with object masks (either given at the first frame or estimated at other frames) as the \textit{memory} frames and the current frame without the object mask as the \textit{query} frame.
The overview of our framework is shown in~\fref{Fig:networks}. 

Both the memory and the query frames are first encoded into pairs of key and value maps through the dedicated deep encoders. 
Note that the query encoder takes only an image as the input, while the memory encoder takes both an image and an object mask. Each encoder outputs \textbf{Key} and \textbf{Value} maps. \textbf{Key} is used for addressing. Specifically, similarities between key features of the query and the memory frames are computed to determine when-and-where to retrieve relevant memory \textbf{values} from. Therefore, \textbf{key} is learned to encode visual semantics for matching robust to appearance variations. 
On the other hand, \textbf{value} stores detailed information for producing the mask estimation (\eg the target object and object boundaries). 
\textbf{Values} from the query and the memory contain information for somewhat different purposes.
Specifically, \textit{value for the query frame} is learned to store detailed appearance information for us to decode accurate object masks. \textit{Value for the memory frames} is learned to encode both the visual semantics and the mask information about whether each feature belongs to the foreground or the background.

The keys and values further go through our space-time memory read block.
Every pixel on the key feature maps of the query and the memory frames is densely matched over the spatio-temporal space of the video.
Relative matching scores are then used to address the value feature map of the memory frame, and the corresponding values are combined to return outputs.
Finally, the decoder takes the output of the read block and reconstructs the mask for the query frame. 


\subsection{Key and Value Embedding}
%
\paragraph{Query encoder.} 
The query encoder takes the query frame as the input.
The encoder outputs two feature maps -- key and value -- through two parallel convolutional layers attached to the backbone network. 
These convolutional layers serve as bottleneck layers to reduce the feature channel size of the backbone network output (to 1/8 for the key and 1/2 for the value) and no non-linearity is applied. 
The output of the query embedding is a pair of 2D key and value maps ($\mathbf{k}^Q\in\mathbb{R}^{H \times W \times C/8}, \mathbf{v}^Q\in\mathbb{R}^{H \times W \times C/2}$), where $H$ is the height, $W$ is the width, and $C$ is the feature dimension of the backbone network output feature map.

\paragraph{Memory encoder.} 
The memory encoder has the same structure except for the inputs.
The input to the memory encoder consists of an RGB frame and the object mask. 
The object mask is represented as a single channel probability map between 0 and 1 (the softmax output is used for estimated masks). 
The inputs are concatenated along the channel dimension before being fed into the memory encoder. 

If there are more than one memory frames, each of them is independently embedded into key and value maps. 
Then, the key and value maps from different memory frames are stacked along the temporal dimension.
The output of the memory embedding is a pair of 3D key and value maps ($\mathbf{k}^M\in\mathbb{R}^{T \times H \times W \times C/8}, \mathbf{v}^M\in\mathbb{R}^{T \times H \times W \times C/2}$), where $T$ is the number of the memory frames.

We take ResNet50~\cite{he2016deep} as the backbone network for both the memory and the query encoder. 
We use the stage-4 (\texttt{res4}) feature map of the ResNet50 as the base feature map for computing the key and value feature maps.
For the memory encoder, the first convolution layer is modified to be able to take a 4-channel tensor by implanting additional single channel filters.
The network weights are initialized from the ImageNet pre-trained model, except for the newly added filters which are initialized randomly. 

\begin{figure}
\centering
\includegraphics[width=1.0\linewidth]{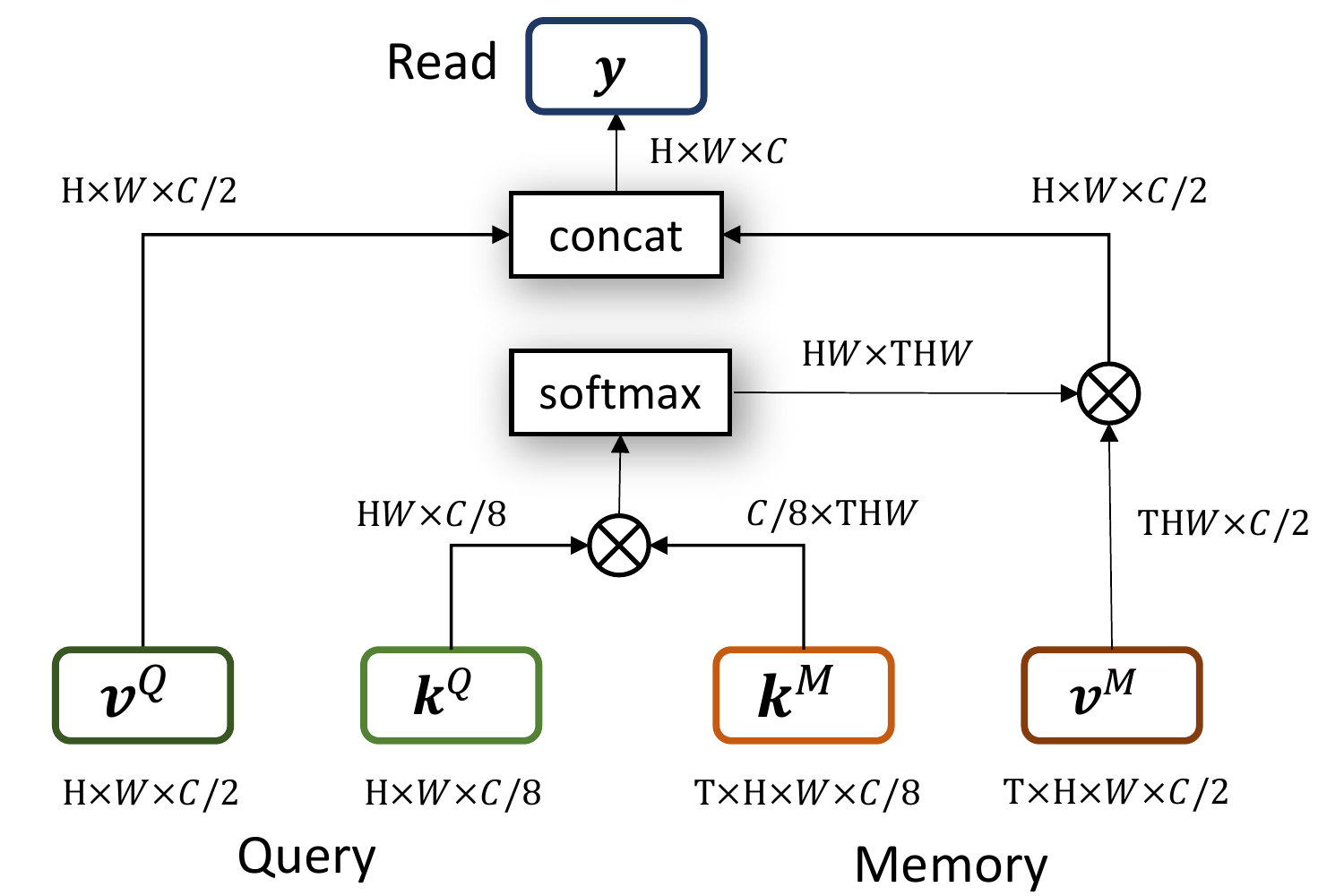}
\caption{Detailed implementation of the space-time memory read operation using basic tensor operations as described in~\sref{read}. $\bigotimes$ denotes matrix inner-product.}
\label{Fig:read}
\end{figure}

\subsection{Space-time Memory Read}\label{read}
In the memory read operation, soft weights are first computed by measuring the similarities between all pixels of the query key map and the memory key map. 
The similarity matching is performed in a non-local manner by comparing every space-time locations in the memory key map with every spatial location in the query key map.
Then, the value of the memory is retrieved by a weighted summation with the soft weights and it is concatenated with the query value. This memory read operates for every location on the query feature map and can be summarized as:  
\begin{equation}
    \mathbf{y}_i = \big[\mathbf{v}^Q_i, \;  \frac{1}{Z}\sum_{\forall j}{f(\mathbf{k}^Q_i,\mathbf{k}^M_j)\mathbf{v}^M_j}\big],
\end{equation}
where $i$ and $j$ are the index of the query and the memory location, $Z = \sum_{\forall j}f(\mathbf{k}^Q_i,\mathbf{k}^M_j)$ is the normalizing factor and $[\cdot,\cdot]$ denotes the concatenation. 
The similarity function $f$ is as follows:
\begin{equation}
    f(\mathbf{k}^Q_i,\mathbf{k}^M_j) = \text{exp}(\mathbf{k}^Q_i\circ\mathbf{k}^M_j),
\label{Eq:similarity}
\end{equation}
where $\circ$ denotes the dot-product.

Our formulation can be seen as an extension of the early formulation of the differential memory networks~\cite{sukhbaatar2015end, miller2016key, kumar2016ask} to 3D spatio-temporal space for video pixel matching.
Accordingly, the proposed read operation localizes the space-time location of the memory for retrieval.
It is also related to non-local self-attention mechanisms~\cite{vaswani2017attention, wang2018non} in that it performs non-local matching. 
However, our formulation is motivated for a different purpose as it is designed to attend to others (memory frames) for the information retrieval, not to itself for the self-attention.
As depicted in~\fref{Fig:read}, our memory read operation can be easily implemented by a combination of basic tensor operations in modern deep learning platforms. 

\subsection{Decoder}
The decoder takes the output of the read operation and reconstructs the current frame's object mask. 
We employ the refinement module used in~\cite{oh2018fast} as the building block of our decoder. 
The read output is first compressed to have 256 channels by a convolutional layer and a residual block~\cite{he2016identity}, then a number of refinement modules gradually upscale the compressed feature map by a factor of two at a time. 
The refinement module at every stage takes both the output of the previous stage and a feature map from the query encoder at the corresponding scale through skip-connections.
The output of the last refinement block is used to reconstruct the object mask through the final convolutional layer followed by a softmax operation. Every convolutional layer in the decoder uses 3$\times$3 filters, producing 256-channel output except for the last one that produces 2-channel output. The decoder estimates the mask in 1/4 scale of the input image.

\subsection{Multi-object Segmentation}
The description of our framework is based on having one target object in the video. 
However, recent benchmarks require a method that can deal with multi-objects~\cite{Pont-Tuset_arXiv_2017, xu2018youtube}. 
To meet this requirement, we extend our framework with a mask merging operation. 
We run our model for each object independently and compute mask probability maps for all objects.
Then, we merge the predicted maps using a soft aggregation operation similar to~\cite{oh2018fast}.
In \cite{oh2018fast}, the mask merging is performed only during the testing as a post-processing step.
In this work, we coin the operation as a differential network layer and apply it during both the training and the testing.
More details are included in the supplementary materials.

\subsection{Two-stage Training}\label{Sect:training}
Our network is first pre-trained on a simulation dataset generated from static image data. 
Then, it is further fine-tuned for real-world videos through the main training. 

\paragraph{Pre-training on images.}
One advantage of our framework is that it does not require long training videos.
This is because the method learns the semantic spatio-temporal matching between distant pixels without any assumption on temporal smoothness. 
This means that we can train our network with only a few frames\footnote{Minimum 2; one as the memory frame and the other as the query.} with object masks. 
This enables us to simulate training videos using image datasets.
Some previous works~\cite{perazzi2017learning, oh2018fast} trained their networks using static images and we take a similar strategy.
A synthetic video clip that consists of 3 frames is generated by applying random affine transforms\footnote{We used rotation, sheering, zooming, translation, and cropping.} to a static image with different parameters.
We leverage the image datasets annotated with object masks (salient object detection -- \cite{shi2016hierarchical, cheng2015global}, semantic segmentation -- \cite{everingham2010pascal, hariharan2011semantic, lin2014microsoft}) to pre-train our network. 
By doing so, we can expect our model to be robust against a wide variety of object appearance and categories. 

\paragraph{Main training on videos.} 
After the pre-training, we use real video data for the main training. 
In this step, either Youtube-VOS~\cite{xu2018youtube} or DAVIS-2017~\cite{Pont-Tuset_arXiv_2017} is used, depending on the target evaluation benchmark.
To make a training sample, we sample 3 temporally ordered frames from a training video. 
To learn the appearance change over a long time, we randomly skip frames during the sampling. 
The maximum number of frames to be skipped is gradually increased from 0 to 25 during the training as in curriculum learning~\cite{yang2015weakly}. 

\paragraph{Training with dynamic memory.}
During the training, the memory is dynamically updated with the network's previous outputs.
As the system moves forward frame-by-frame, the computed segmentation output at the previous step is added to the memory for the next frame.
The raw network output without thresholding, which is a probability map of being a foreground object, is directly used for the memory embedding to model the uncertainty of the estimation.

\paragraph{Training details.}
We used randomly cropped 384$\times$384 patches for training. 
For all experiments, we set the mini-batch size to 4 and disabled all the batch normalization layers.  
We minimize the cross-entropy loss using Adam optimizer~\cite{kingma2014adam} with a fixed learning rate of 1e-5. 
Pre-training takes about 4 days and main training takes about 3 days using four NVIDIA GeForce 1080 Ti GPUs.

\subsection{Inference}\label{Sect:inference}
%
Writing all previous frames on to the memory may raise practical issues such as GPU memory overflow and slow running speed. 
Instead, we select frames to be put onto the memory by a simple rule.  
The first and the previous frame with object masks are the most important reference information~\cite{perazzi2017learning, oh2018fast, yang2018efficient}. 
The first frame always provides reliable information as it comes with the ground truth mask. 
The previous frame is similar in appearance to the current frame, thus we can expect accurate pixel matching and mask propagation. 
Therefore, we put these two frames into the memory by default.

For the intermediate frames, we simply save a new memory frame every $N$ frames. 
$N$ is a hyperparameter that controls the trade-off between speed and accuracy, and we use $N=5$ unless mentioned otherwise.  
It is noteworthy that our framework achieves the effect of online learning and online adaptation without additional training. 
The effect of online model updating is easily accomplished by putting the previous frames into the memory without updating model parameters. 
Thus, our method runs considerably faster than most of the previous methods while achieving state-of-the-art accuracy. 

\section{Evaluation}
We evaluate our model on Youtube-VOS~\cite{xu2018youtube_tech} and DAVIS~\cite{Perazzi2016,Pont-Tuset_arXiv_2017} benchmarks. 
We prepared two models trained on each benchmarks' training set.
For the evaluation on Youtube-VOS, we used 3471 training videos following the official split~\cite{xu2018youtube_tech}.
For DAVIS, we used 60 videos from the DAVIS-2017 train set.
Both DAVIS 2016 and 2017 are evaluated using a single model trained on DAVIS-2017 for a fair comparison with the previous works~\cite{oh2018fast, yang2018efficient}.
In addition, we provide the results for the DAVIS with our model trained with additional training data from Youtube-VOS.
Note that we use the network output directly without post-processing to evaluate our method.

We measured region similarity $\mathcal{J}$ and contour accuracy $\mathcal{F}$. 
For Youtube-VOS, we uploaded our results to the online evaluation server~\cite{xu2018youtube_tech}.
For DAVIS, we used the official benchmark code~\cite{Perazzi2016}. 
Our code and model will be available online.

\begin{table}
\centering 
\begin{tabular}{lccccc}
\toprule
 & & \multicolumn{2}{c}{Seen} & \multicolumn{2}{c}{Unseen} \\
\cmidrule(lr){3-4}
\cmidrule(lr){5-6}
&  Overall & $\mathcal{J}$ & $\mathcal{F}$ & $\mathcal{J}$  & $\mathcal{F}$\\
\midrule
OSMN~\cite{yang2018efficient} & 51.2 & 60.0 & 60.1 & 40.6 & 44.0 \\
MSK~\cite{perazzi2017learning} & 53.1 & 59.9 & 59.5 & 45.0 & 47.9 \\
RGMP~\cite{oh2018fast} & 53.8 & 59.5 & - & 45.2 & - \\
OnAVOS~\cite{voigtlaender2017online} & 55.2 & 60.1 & 62.7 & 46.6 & 51.4 \\
RVOS~\cite{rvos2019} & 56.8 & 63.6 & 67.2 & 45.5 & 51.0 \\
OSVOS~\cite{caelles2017one} & 58.8 & 59.8 & 60.5 & 54.2 & 60.7 \\
S2S~\cite{xu2018youtube} & 64.4 & 71.0 & 70.0 & 55.5 & 61.2 \\
A-GAME~\cite{joakim2018generative} & 66.1 & 67.8 & - & 60.8 & - \\
PreMVOS~\cite{luiten2018premvos} & 66.9 & 71.4 & 75.9 & 56.5 & 63.7 \\
BoLTVOS~\cite{voigtlaender2019boltvos}  & 71.1 & 71.6 & - & 64.3 & - \\
\midrule
Ours & \textbf{79.4} & \textbf{79.7} & \textbf{84.2} &  \textbf{72.8} & \textbf{80.9} \\

\bottomrule
\end{tabular}
\caption{The quantitative evaluation of multi-object video object segmentation on Youtube-VOS~\cite{xu2018youtube} validation set. Results for other methods are directly copied from~\cite{xu2018youtube_tech, joakim2018generative, rvos2019, voigtlaender2019boltvos}.}
\label{Table:ytvos}
\end{table}

\begin{table}
\centering 
\begin{tabular}{p{2.6cm}cccc}
\toprule
 & OL & $\mathcal{J}$ Mean  & $\mathcal{F}$ Mean  & Time\\
\midrule
S2S \textbf{(+YV)}~\cite{xu2018youtube} & \checkmark & 79.1 & - & 9$s$ \\
MSK~\cite{perazzi2017learning}   & \checkmark & 79.7 & 75.4  & 12$s$\\ 
OSVOS~\cite{caelles2017one}  & \checkmark & 79.8 & 80.6  & 9$s$\\
MaskRNN~\cite{hu2017maskrnn} &  \checkmark & 80.7 & 80.9  & -\\
\multicolumn{2}{l}{VideoMatch~\cite{hu2018videomatch}} &  81.0 & - &0.32$s$ \\
\multicolumn{2}{l}{FEELVOS \textbf{(+YV)}~\cite{feelvos2019}} &  81.1 & 82.2 & 0.45$s$\\
RGMP~\cite{oh2018fast}&  & 81.5 & 82.0  &  0.13$s$ \\
\multicolumn{2}{l}{A-GAME \textbf{(+YV)}~\cite{joakim2018generative}} &  82.0 & 82.2 & 0.07$s$\\
FAVOS~\cite{cheng2018fast} &  & 82.4 & 79.5  & 1.8$s$ \\
LSE~\cite{ci2018video}& \checkmark   &  82.9 & 80.3 & - \\
CINN~\cite{bao2018cnn}  & \checkmark & 83.4 & 85.0 &  $>$30$s$ \\
PReMVOS~\cite{luiten2018premvos}& \checkmark & 84.9 & 88.6  & $>$30$s$ \\
OSVOS$^S$~\cite{maninis2017video}   & \checkmark &  85.6 & 86.4 & 4.5$s$\\
OnAVOS~\cite{voigtlaender2017online}  & \checkmark &  86.1 & 84.9  & 13$s$\\ 
DyeNet~\cite{li2018video} & \checkmark & 86.2 & -   &  2.32$s$ \\
\midrule
Ours &   & 84.8 & 88.1 & 0.16$s$ \\ 
Ours \textbf{(+YV)} &   & \textbf{88.7} & \textbf{89.9} & 0.16$s$ \\ 
\bottomrule
\end{tabular}
\caption{The quantitative evaluation on DAVIS-2016 validation set. 
OL indicates online learning. 
\textbf{(+YV)} indicates the use of Youtube-VOS for training. 
Methods with $\mathcal{J}$ Mean below 79 are omitted due to the space limit and the complete table is available in the supplementary material.
}
\label{Table:DAVIS2016}
\end{table}

\begin{table}
\centering
\begin{tabular}{lccc}
\toprule
 & OL & $\mathcal{J}$ Mean & $\mathcal{F}$ Mean \\ 
\midrule
OSMN~\cite{yang2018efficient} & & 52.5 & 57.1  \\
FAVOS~\cite{cheng2018fast} & & 54.6 & 61.8 \\
VidMatch~\cite{hu2018videomatch} & & 56.5 & 68.2 \\
OSVOS~\cite{caelles2017one}  & \checkmark & 56.6 & 63.9  \\
MaskRNN~\cite{hu2017maskrnn}  & \checkmark & 60.5 & -  \\
OnAVOS~\cite{voigtlaender2017online} & \checkmark & 64.5 & 71.2 \\ 
OSVOS$^S$~\cite{caelles2017one} & \checkmark & 64.7 & 71.3  \\
RGMP~\cite{oh2018fast}  & & 64.8 & 68.6 \\
CINN~\cite{bao2018cnn} & \checkmark & 67.2 & 74.2 \\
A-GAME \textbf{(+YV)}~\cite{joakim2018generative} & & 67.2 & 72.7 \\
FEELVOS \textbf{(+YV)}~\cite{feelvos2019} & & 69.1 & 74.0 \\
DyeNet~\cite{li2018video} & \checkmark & *74.1 & \\
PReMVOS~\cite{luiten2018premvos} & \checkmark & 73.9 & 81.7  \\
\midrule
Ours & & 69.2 & 74.0 \\ 
Ours \textbf{(+YV)} & & \textbf{79.2} & \textbf{84.3} \\ 
\bottomrule
\end{tabular}
\caption{The quantitative evaluation on DAVIS-2017 validation set.
OL indicates online learning. 
\textbf{(+YV)} indicates the use of Youtube-VOS for training. *: average of $\mathcal{J}$ Mean and $\mathcal{F}$ Mean.}
\label{Table:DAVIS2017}
\end{table}

\subsection{Youtube-VOS} 
Youtube-VOS~\cite{xu2018youtube} is the latest large-scale dataset for the video object segmentation that consists of 4453 videos annotated with multiple objects.
The dataset is about 30 times larger than the popular DAVIS benchmark that consists of 120 videos. 
It also has validation data for the unseen object categories.
Thus, it is good for measuring the generalization performance of different algorithms. 
The validation set consists of 474 videos including 91 object categories. 
It has separate measures for 65 of seen and 26 of unseen object categories. 
We compare our method to existing methods that are trained on Youtube-VOS training set by~\cite{joakim2018generative, xu2018youtube_tech}.
As shown in~\Tref{Table:ytvos}, our method significantly outperforms all other methods in every evaluation metric. 


\subsection{DAVIS} 
\paragraph{Single object (DAVIS-2016).} 
DAVIS-2016~\cite{Perazzi2016} is one of the most popular benchmark datasets for video object segmentation tasks. We use the validation set that contains 20 videos annotated with high-quality masks each for a single target object. 
We compare our method with state-of-the-art methods in \Tref{Table:DAVIS2016}. 
In the table, we indicate the use of online learning and provide approximate runtimes of each method. 
Most of the previous top-performing methods rely on online learning that severely harms the running speed.
Our method achieves the best accuracy among all competing methods without online learning, and shows competitive results with the top-performing online learning based methods while running in a fraction of time. 
Our method trained with additional data from Youtube-VOS outperforms all the methods by a large margin.

\begin{figure*}
\centering
\includegraphics[width=1.0\linewidth]{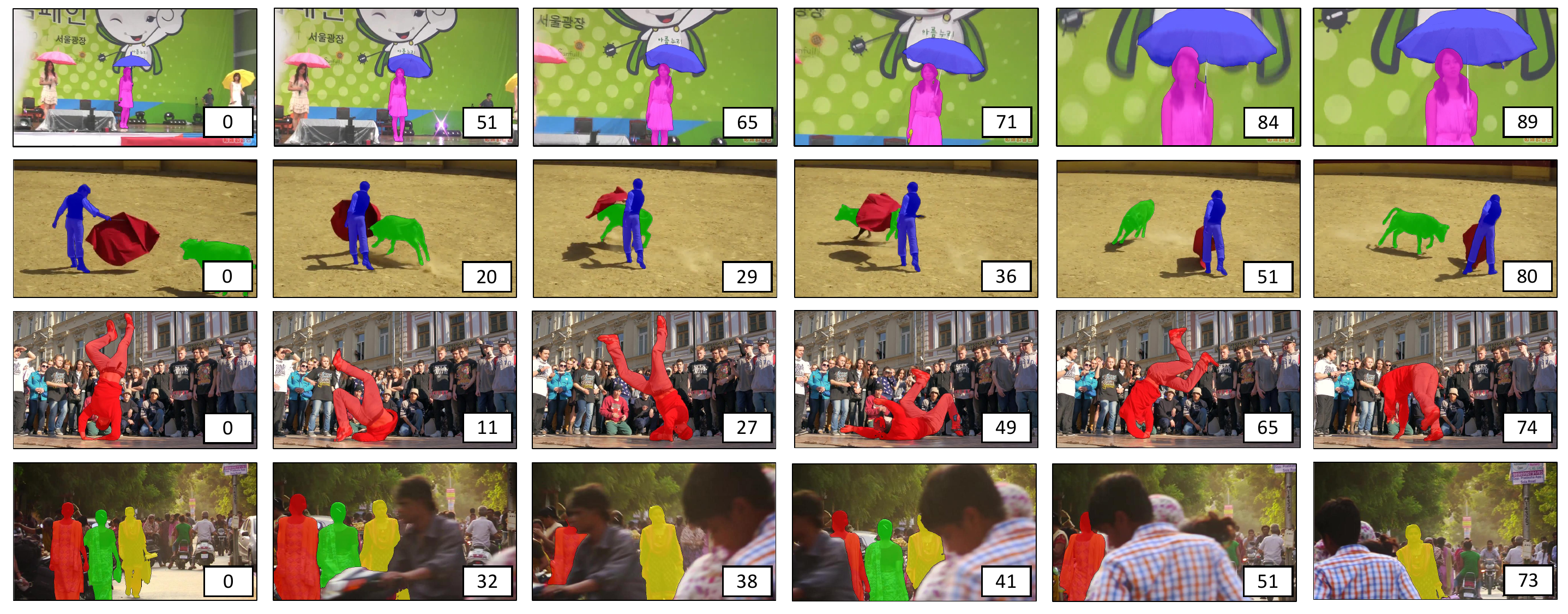}
\caption{The qualitative results on Youtube-VOS and DAVIS. Frames are sampled at important moments (\eg before and after occlusions).}
\label{Fig:qualitative}
\end{figure*}

\paragraph{Multiple objects (DAVIS-2017).}  
DAVIS-2017~\cite{Pont-Tuset_arXiv_2017} is a multi-object extension of DAVIS-2016. 
The validation set consists of 59 objects in 30 videos.
In Table~\Tref{Table:DAVIS2017}, we report the results of multi-object video segmentation on the validation set. 
Again, our method shows the best performance among fast methods without online learning.
With additional Youtube-VOS data, our method largely outperforms all the previous state-of-the-art methods including the 2018 DAVIS challenge winner~\cite{luiten2018premvos}. 
Our results on the test-dev set is included in the supplementary materials.

The large performance leap by using additional training data indicates that DAVIS is too small to train a generalizable deep network due to over-fitting. 
It also explains why top performing online learning methods on the DAVIS benchmark do not show good performance on the large-scale Youtube-VOS benchmark. 
Online learning methods are hardly aided by large training data. 
Those methods usually require an extensive parameter search (\eg data synthesis methods, optimization iterations, learning rate, and post-processing), which is not easy to do for a large-scale benchmark.

\subsection{Qualitative Results.} 
\fref{Fig:qualitative} shows qualitative examples of our results. We choose challenging videos from Youtube-VOS and DAVIS validation sets and sample important frames (\eg before and after occlusions). 
As can be seen in the figure, our method is robust to occlusions and complex motions.
More results will be provided in the supplementary material.

\begin{figure*}
\centering
\includegraphics[width=1.0\linewidth]{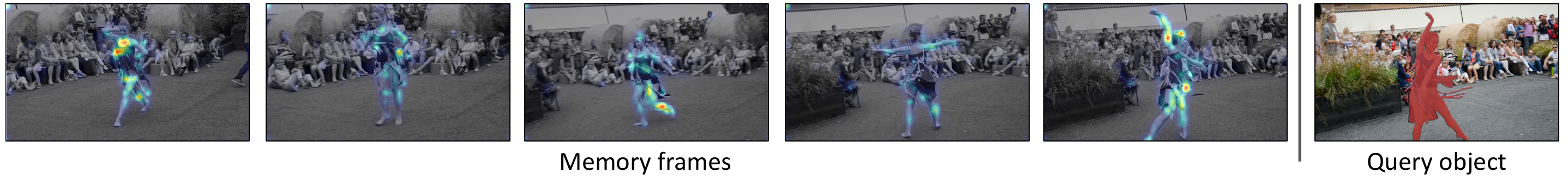}
\caption{Visualization of our space-time read operation. We first compute the similarity scores in \eref{Eq:similarity} for the pixels inside the object area of the query image (marked in red), then visualize the normalized soft weights to the memory frames.
}
\label{Fig:viz}
\end{figure*}

\begin{table}
\centering 
\begin{tabular}{lc|P{1cm}P{1cm}}
\toprule
Variants & Youtube-VOS & \multicolumn{2}{c}{DAVIS-2017} \\
\cmidrule(lr){2-2}
\cmidrule(lr){3-4}
 & Overall & $\mathcal{J}$ & $\mathcal{F}$\\
\midrule
Pre-training only & 69.1  & 57.9 & 62.1 \\
Main-training only &  68.2 &  38.1 & 47.9 \\
Full training & \textbf{79.4}  &  69.2 & 74.0 \\
\midrule
Cross validation & 56.3  &  \textbf{78.6} & \textbf{83.5} \\
\bottomrule
\end{tabular}
\caption{Training data analysis on Youtube-VOS and DAVIS-2017 validation sets. We compare models trained through different training stages (\sref{Sect:training}). In addition, we report the cross-validation results (\ie evaluating DAVIS using the model trained on Youtube-VOS, and vice versa.).}
\label{Table:training}
\end{table}



\begin{table}
\centering 
\begin{tabular}{lc|P{0.95cm}P{0.95cm}|c}
\toprule
\multirowcell{2}[0ex][l]{Memory\\frame(s)} & \multirowcell{2}[-0.4ex][c]{Youtube\\-VOS} & \multicolumn{2}{c|}{DAVIS} &  \multirowcell{2}[-0.4ex][c]{Time}\\

\cmidrule(lr){3-4}
&  & 2016  & 2017 & \\
\midrule
First & 68.9   & 81.4 & 67.0 & 0.06 \\
Previous &  69.7 &  83.2 & 69.6 & 0.06 \\
First \& Previous & 78.4  &  87.8 & 77.7 & 0.07  \\
Every 5 frames & \textbf{79.4}  &  \textbf{88.7} & \textbf{79.2} & 0.16 \\
\bottomrule
\end{tabular}
\caption{Memory management analysis on the validation sets of Youtube-VOS and DAVIS. We compare results obtained by different memory management rules. We report \textit{Overall} and $\mathcal{J}$ Mean scores for Youtube-VOS and DAVIS, respectively. Time is measured on DAVIS-2016.}
\label{Table:memory}
\end{table}

\section{Analysis}
\subsection{Training Data} \label{TD}
We trained our model through two training stages: the pre-training on static images~\cite{shi2016hierarchical, cheng2015global,everingham2010pascal, hariharan2011semantic, lin2014microsoft} and the main training using DAVIS~\cite{Pont-Tuset_arXiv_2017} or Youtube-VOS~\cite{xu2018youtube}.
In~\Tref{Table:training}, we compare the performance of our method with different training data. 
In addition, we provide a cross-dataset validation to measure the generalization performance.

\paragraph{Pre-training only.} 
It is interesting that our pre-train only model outperforms the main-train only model as well as all other methods on YouTube-VOS, without using any real video.
However, we get maximum performance by using both training strategies.

\paragraph{Main-training only.} 
Without the pre-training stage, the performance of our model drops by 11.2 in \textit{Overall} score on Youtube-VOS~\cite{xu2018youtube}.
This indicates that the amount of training video data is still not enough to bring out the potential of our network even though Youtube-VOS~\cite{xu2018youtube} provides more than 3000 training videos. 
In addition, very low performance on DAVIS implies a severe over-fitting issue as the training loss was similar to the complete model (We did not apply early stopping). 
We conjecture that diverse objects encountered during the pre-training helped our model's generalization and also to prevent over-fitting.  

\paragraph{Cross validation.}
We evaluate our model trained on DAVIS to Youtube-VOS, and vice versa. 
Our model trained on DAVIS shows limited performance on Youtube-VOS.
This is an expected result because DAVIS is too small to learn a generalization ability to other datasets.
On the other hand, our model trained on Youtube-VOS performs well on DAVIS and outperforms all other methods. 

\begin{figure}
\centering
\includegraphics[width=1.0\linewidth]{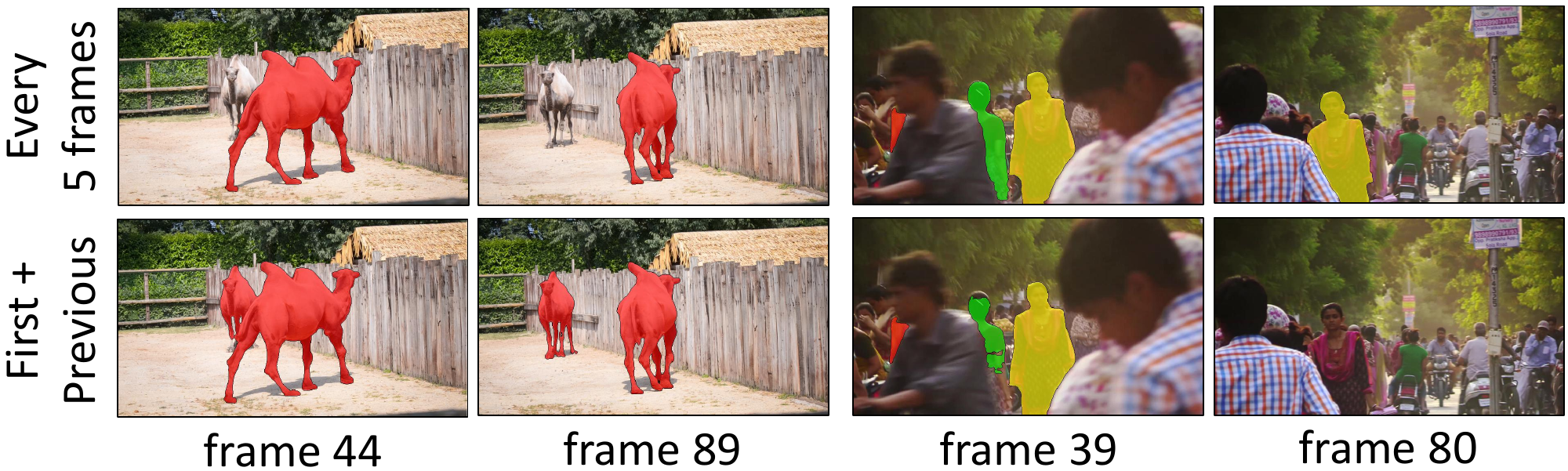}
\caption{Visual comparisons of the results with and without using the intermediate frame memories.}
\label{Fig:vs_FL}
\end{figure}

\subsection{Memory Management} \label{MM}
For the minimal memory consumption and fastest runtime, we can save either the first and/or the previous frames in the memory. For the maximum accuracy, our final model saves a new intermediate memory frame at every 5 frames in addition to the first and the previous frames as explained in \sref{Sect:inference}.

We compare different memory management rules in~\Tref{Table:memory}. 
Saving both the first and the previous frame into the memory is the most important, and our model achieves state-of-the-art accuracy with the two memory frames. This is because our model is strong enough to handle large appearance changes while being robust to drifting and error accumulation by effectively exploiting the memory.
On top of that, having the intermediate frame memories further boosts performance by tackling extremely challenging cases as shown in~\fref{Fig:vs_FL}.

\begin{figure}
\centering
\includegraphics[width=1.0\linewidth]{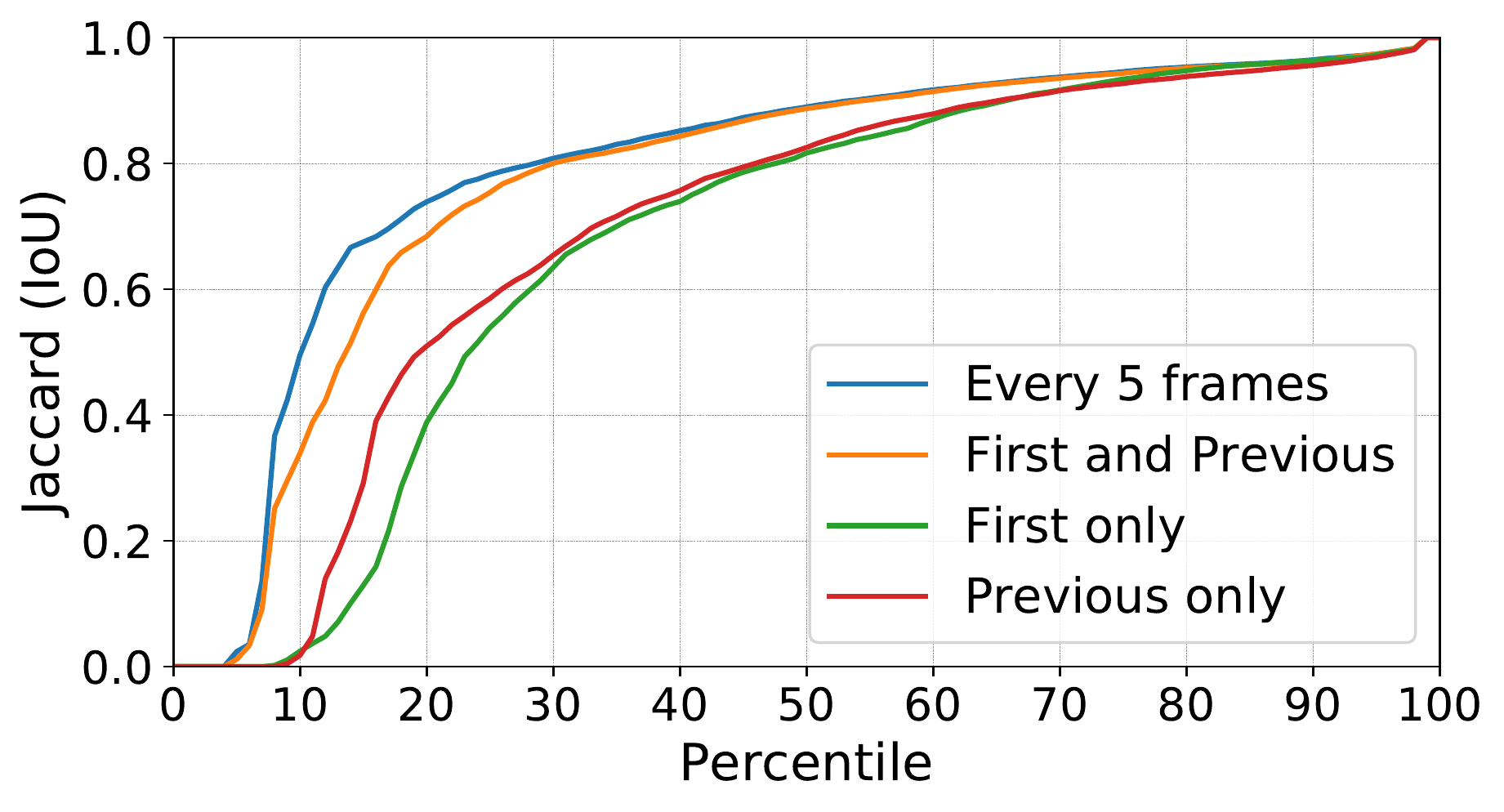}
\vspace{-5mm}
\caption{Jaccard score distribution on DAVIS-2017.}
\label{Fig:robust}
\end{figure}

For a deeper analysis, we show the frame-level accuracy distribution in~\fref{Fig:robust}. 
We sort Jaccard scores of all objects in all video frames and plot the scores to analyze the performance on challenging scenes. 
We compare our final model (\textit{Every 5 frames}) with \textit{First and Previous} to check the effect of using additional memory frames.
While both settings perform equally well on the successful range (over 30$^\text{th}$ percentile), the effect of additional memory frames becomes clearly visible for difficult cases (under 30$^\text{th}$ percentile). 
The huge accuracy gap between 10 and 30 percentile indicates that our network handles challenging cases better with additional memory frames.
Comparing \textit{First only} and \textit{Previous only}, the previous frame looks more useful to handle failure cases. 

\paragraph{Memory visualization.}
In~\fref{Fig:viz}, we visualize our memory read operation to validate the learned space-time matching. 
As can be observed in the visualization, our read operation accurately matches corresponding pixels between the query and the memory frames.

\section{Conclusion}
We have presented a novel space-time memory network for the semi-supervised video object segmentation. 
Our method performs the best among the existing methods in terms of both the accuracy and the speed. 
We believe the proposed space-time memory network has a great potential to become breakthroughs in many other pixel-level estimation problems. 
We are looking for other applications as future work that are suited for our framework including object tracking, interactive image/video segmentation, and inpainting.   \\

\small{
\noindent\textbf{Acknowledgment.} This work is supported by the ICT R\&D program of MSIT/IITP (2017-0-01772, Development of QA systems for Video Story Understanding to pass the Video Turing Test).
}

{\small
\bibliographystyle{ieee_fullname}
\bibliography{egbib}
}

\end{document}